\documentclass[runningheads]{llncs}

\usepackage{eccv}

\usepackage{eccvabbrv}

\usepackage{graphicx}
\usepackage{booktabs}

\usepackage[accsupp]{axessibility}  %

\usepackage{hyperref}

\usepackage{orcidlink}

\usepackage{algorithm2e}
\usepackage{algpseudocode}
\usepackage[inline]{enumitem}

\begin{document}

\title{Model Breadcrumbs: Scaling Multi-Task Model Merging with Sparse Masks} 

\titlerunning{Model Breadcrumbs}

\author{MohammadReza Davari\and Eugene Belilovsky}

\authorrunning{MR. Davari and E. Belilovsky}

\institute{Concordia University and Mila -- Quebec AI Institute
\email{\{mohammadreza.davari,eugene.belilovsky\}@concordia.ca}}

\maketitle

\RestyleAlgo{ruled}
\SetKwComment{Comment}{/* }{ */}

\begin{abstract}
  The rapid development of AI systems has been greatly influenced by the emergence of foundation models. A common approach for targeted problems involves fine-tuning these pre-trained foundation models for specific target tasks, resulting in a rapid spread of models fine-tuned across a diverse array of tasks. This work focuses on the problem of merging multiple fine-tunings of the same foundation model derived from a spectrum of auxiliary tasks. We introduce a new simple method, Model Breadcrumbs, which consists of a sparsely defined weight set that guides model adaptation within the weight space of a pre-trained model. These breadcrumbs are constructed by subtracting the weights from a pre-trained model before and after fine-tuning, followed by a sparsification process that eliminates weight outliers and negligible perturbations. Our experiments demonstrate the effectiveness of Model Breadcrumbs to simultaneously improve performance across multiple tasks. This contribution aligns with the evolving paradigm of updatable machine learning, reminiscent of the collaborative principles underlying open-source software development, fostering a community-driven effort to reliably update machine learning models. Our method is shown to be more efficient and unlike previous proposals does not require hyperparameter tuning for each new task added. Through extensive experimentation involving various models, tasks, and modalities we establish that integrating Model Breadcrumbs offers a simple, efficient, and highly effective approach for constructing multi-task models and facilitating updates to foundation models.
  \footnote{The code to reproduce our results is publicly available at:  \href{https://github.com/rezazzr/breadcrumbs}{https://github.com/rezazzr/breadcrumbs}}
\end{abstract}

\section{Introduction}

In recent years, foundational models~\cite{bommasani2021opportunities} have become instrumental tools, exhibiting unprecedented efficacy across multiple domains. These models are characterized by their extensive scale, generality, and capacity to learn and generalize knowledge from vast datasets, offering promising solutions to a diverse range of problems. The inherent ability of foundational models to be fine-tuned has led to advancements in natural language processing (NLP)~\cite{radford2018gpt,radford2019language-gpt2,devlin2018bert,liu2019roberta,raffel2020exploring-t5,lewis2019bart}, computer vision~\cite{radford2021learning,ramesh2021zero-dalle,luo2020univl,kim2021vilt,cho2021unifying-vision-text}, and other related fields~\cite{rives2021biological-250mil,yin2020tabert,rothchild2021c5t5}. 

On one hand, the scalability of expanding foundational models to increase the number of tasks they can perform in practice poses a significant challenge as approaches such as joint training are limited in many practical scenarios~\cite{cossu2022continual,davari2022probing}. In domains such as healthcare, stringent data privacy concerns often prohibit access to the underlying training data, even when the fine-tuned model on the said data is publicly accessible, rendering joint training infeasible~\cite{asadi2023prototype,davari2021probing}. Even in scenarios where access to training data is possible, the computational demands of simultaneous training on a multitude of tasks becomes restraining.

On the other hand, the widespread adoption of foundational models has led to a certain homogenization in the field~\cite{bommasani2021opportunities}. Both the training approach, commonly transfer learning from a popular foundational model~\cite{oquab2014learning}, and the model architecture itself have become standardized, typically following a few popular foundation models. This standardization has resulted in a proliferation of publicly available fine-tuned models, all sharing the same architecture~\cite{wolf2020transformers, yang2023toxbuster,davari2020timbert, farahnak2021semantic}. However, beyond their conventional use for model inference, these numerous fine-tuned models remain \textbf{largely untapped}, representing a missed opportunity~\cite{rame2023ratatouille}.

To address the challenges of scalability, practical constraints, and unlock the untapped potential of the growing pool of publicly available fine-tuned models, recent developments in neural network weight averaging techniques have gained attention~\cite{izmailov2018averaging, davari2024model, neyshabur2020being, rame2023ratatouille, ilharco2022editing, wortsman2022robust, wortsman2022model,choshen2022fusing, don2022cold,yadav2023ties}. These approaches enable the practitioners to re-purpose the increasingly valuable publicly available fine-tuned models. 

Closer to our approach, \textit{Task Arithmetic} were introduced by Ilharco \etal~\cite{ilharco2022editing}. In their method, a foundation model is refined by incorporating the scaled average of the differences between multiple fine-tuned models and the foundation model. This allows for the creation of a multi-task model without the need for additional training or access to the original training data. However, despite its potential, the Task Arithmetic method~\cite{ilharco2022editing} encounters limitations when dealing with numerous tasks. This is mainly due to its dependence on hyperparameter tuning through validation set performance, a process that becomes computationally impractical at scale, coupled with an increasing accumulation of noise as more tasks are merged to the foundation model.

\begin{figure}[t]
    \centering
    \includegraphics[width=0.95\textwidth]{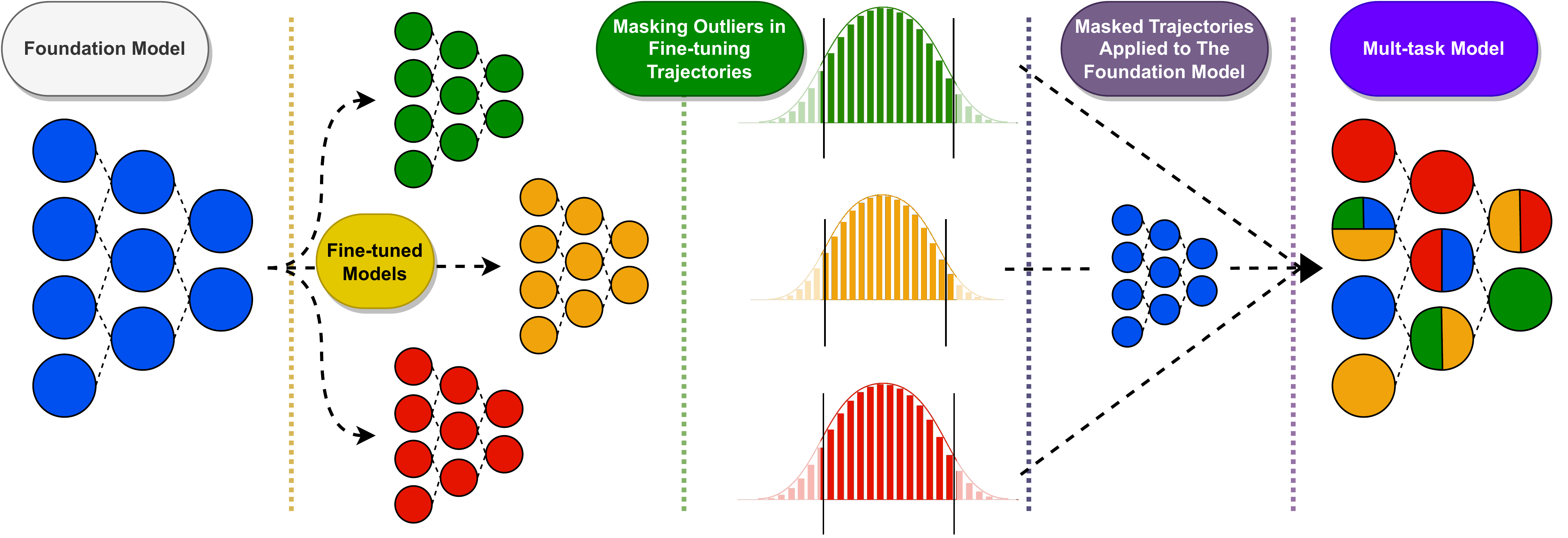}
    \caption{Method overview. We start with a foundational model that has undergone fine-tuning on various tasks. Next, we build a fine-tuning trajectory for each fine-tuned model by subtracting the pre-trained model weights from each of the fine-tuned models (task vectors). We then, at each layer, apply a masking operation over the absolute value of the the resulting trajectory, eliminating both outliers and small values. Finally, these masked task vectors are aggregated and combined with the reference pre-trained model to create a unified multi-task model.}
    \label{fig:method-overview}
\end{figure}

To address these challenges and to capitalize on the untapped resources within the field, our paper introduces Model Breadcrumbs, a simple solution designed to tackle scalability, noise reduction in merging tasks, and hyperparameter generalization issues. Model Breadcrumbs constructs multi-task models from pre-existing fine-tuned models (see Figure~\ref{fig:method-overview}), surpassing limitations faced by existing methods. We demonstrate that Model Breadcrumbs not only yields competitive multi-task models but also provides hyperparameters that generalize effectively as the number of tasks increases. In Section~\ref{sec:related-work}, we provide context through a review of related work. Sections~\ref{sec:framework} and \ref{sec:exp} present our framework and its evaluation. Finally, Section~\ref{sec:conclusion} outlines the scope and limitations of our proposed method. Our key contributions and findings are summarized as follows:

\begin{enumerate}
    \item Introducing a simple and scalable approach for merging models and reusing pre-existing fine-tuned models to construct multi-task models, often outperforming their individual fine-tuned counterparts.
    \item We empirically show the robustness of our approach to hyperparameter variations and its ability to generalize with the increasing number of tasks.
\end{enumerate}

\section{Related Work}
\label{sec:related-work}

\noindent\textbf{Model Merging}\quad
Recent studies in the literature have explored the merging of models trained from scratch with different initializations~\cite{ainsworth2022git,stoica2023zipit}. One of the main challenges in this type of model merging is aligning the models before the actual merger. Therefore, research in this branch primarily focuses on finding permutations between networks to bring them into alignment with a reference model, enabling the subsequent merger of the two models in weight space.
Our work, on the other hand, distinguishes itself from this line of research, as we concentrate on the model merging of networks that share the same initialization, specifically initialized by a foundation model. Furthermore, our investigation is focused on the scalability of merging methods, exploring the dynamics when multiple models are involved in the merger process.

Neyshabur \etal~\cite{neyshabur2020being} highlighted the benefits of linearly interpolating two fine-tuned models originating from the same pre-trained model. They showed that this technique often yields a model that outperforms both of the original fine-tuned models. This discovery sparked subsequent investigations into the merging of fine-tuned models derived from a single foundation model, exploring its potential and practical applications.

Wortsman \etal~\cite{wortsman2022model} demonstrated that models fine-tuned on the same dataset with different hyperparameters can be combined together using a weighted average to yield an overall higher performing model. Unlike our work they did not consider merging models from different datasets and tasks. Choshen \etal~\cite{choshen2022fusing} merged models from multiple trained models in order to create a better pretrained model to be used for downstream tasks. Unlike our work they do not demonstrate or study the creation of multi-task ability through the merging. 
Matena and Raffel~\cite{NEURIPS2022_70c26937} considered merging of multiple fine-tuned models originating from the same pre-trained model, trained on diverse datasets. The merger operation combines a series of fine-tuned models using a weighted average determined by the Fisher information matrix~\cite{myung2003tutorial}. However, computing the Fisher information matrix, as well as finding other required hyperparameters for this approach, becomes increasingly computationally expensive as the number of models to be merged grows. Therefore, it faces challenges when applied at scale. In contrast, our approach is computationally efficient, and as we will show in Section~\ref{sec:exp}, its hyperparameters exhibit the ability to generalize to the scenarios where a large number of models are to be merged. 

A related study to ours is conducted by Ilharco \etal~\cite{ilharco2022editing}, introducing a method named Task Arithmetic for model merging. Their approach begins by forming \textit{Task Vectors}, representing the weight differences between pre-trained and fine-tuned weights for each task. The merged model's weights are then obtained by adding a scaled sum of these task vectors to the pre-trained weights.
However, their approach necessitates a validation set for each new task, which adds complexity and computational overhead, coupled with an increasing accumulation of noise as more tasks are merged to the foundation model.

A concurrent study by Yadav \etal~\cite{yadav2023ties} presents a method named TIES. Like the Task Arithmetic method~\cite{ilharco2022editing}, TIES initially constructs a set of Task Vectors. These vectors undergo a masking process to eliminate interfering weights, identified as a percentage of overall weights with low magnitudes. The remaining unmasked weights undergo a sign alignment operation to determine their polarity. Finally, a scaled sum merges the task vectors with the pre-trained model.
Our approach differs from TIES in two key aspects. Firstly, we apply masking to both very large and small magnitude weights of the task vectors to minimize interference, whereas TIES focuses solely on small magnitude weights. Secondly, our masking strategy employs layer-wise masking as opposed to overall masking. Notably, in the context of task vectors, overall masking of small magnitude weights typically targets weights in the early layers~\cite{NEURIPS2022_70c26937}. 

\noindent\textbf{Federated Learning}\quad
The concept of initiating learning with a pre-trained model has been explored in the federated learning literature, as seen in recent works such as \cite{nguyen2022begin,legate2023guiding}. These studies focused on a single downstream task where data is distributed across multiple clients. In their approach, each client periodically aggregates models during the training process. It's important to note that this differs from our approach, which deals with multi-task learning involving multiple downstream tasks rather than a single task distributed across clients.

\section{Model Breadcrumbs Framework}
\label{sec:framework}

The Model Breadcrumbs framework is designed to enable the construction of multi-task models from pre-existing fine-tuned foundation models without the need for further training. The central idea is to merge models and aggregate valuable \textit{knowledge} for the resulting multi-task model while filtering out potential harmful perturbations that could impact its performance. This section provides an overview of the process for acquiring and merging Model Breadcrumbs.

To start generating Model Breadcrumbs, we begin with a pre-trained foundation model that has undergone fine-tuning for various auxiliary tasks. Denoting the weights of the foundation model as $\theta$, after fine-tuning on a specific task $t$, the weights are transformed into $\theta'_t$. The initial step involves creating \textit{task vectors} by calculating the weight differences between $\theta'_t$ and $\theta$, resulting in $\theta^d_t$.

\begin{equation}
    \theta^d_t = \theta'_t - \theta
\end{equation}

Note that $\theta^d_t$ contains both (a) large outliers, indicating substantial deviations from the pre-trained starting point, and (b) negligible differences representing minor perturbations from the foundation model's weights. The presence of these extremes can impact the effectiveness of the resulting multi-task model upon merging. To address this concern, we implement a masking process that filters out both large outliers and small perturbations.

In each layer $L$, we mask out the extreme tails of the absolute magnitude distribution of $\theta^d_t$, using $\gamma$ and $\beta$ as thresholds for the right and left tails, respectively. Let $w^L_i$ represent the index of the weights sorted by their absolute magnitude in layer $L$ and $i$ the order in the sort (lowest to highest). The mask $m_L^{\beta,\gamma}$ for the layer $L$ is defined as:

\begin{equation} 
 m_L^{\beta,\gamma}[w_i]=
         \begin{cases}
         0 & \text{if } i \leq \beta \text{ or } i \geq \gamma \\
    1 & \text{otherwise}
        \end{cases} 
\end{equation}

The masked weights are set to zero in $\theta^d_t$ or returned to their respective pre-training weights in $\theta'_t$. Aggregating $m_L^{\beta,\gamma}$ over all layers, for task $t$, results in the final mask $m^{\beta, \gamma}_t$.

Next, we apply the mask  $m^{\beta, \gamma}_t$ to the task vectors $\theta^d_t$. We now have a set of weight differences that define a trajectory within the weight space of the foundation model. Traversing this trajectory allows us to effectively transfer the knowledge accumulated during fine-tuning across tasks, while filtering out the harmful perturbations. For a total of $T$ tasks, we assemble a multi-task model $\theta^*$ by following the trajectories defined by the Model Breadcrumbs with a specific strength parameter $\alpha$. The formation of this multi-task model is expressed in Eq.~\ref{eq:multi-task-model}. Algorithm~\ref{alg:breadcrumbs} describes the overall procedure of the Model Breadcrumbs merging strategy.

\begin{equation}
    \theta^* = \theta + \alpha \sum_{t \in T} m^{\beta, \gamma}_t .  \theta^d_t
    \label{eq:multi-task-model}
\end{equation}

\makeatletter
\patchcmd{\algocf@makecaption@ruled}{\hsize}{\textwidth}{}{} %
\patchcmd{\@algocf@start}{-2.5em}{0em}{}{} %
\makeatother
\setlength{\algomargin}{0em}

\begin{algorithm}[t]
\caption{Model merging via Breadcrumbs.}\label{alg:breadcrumbs}
\KwData{Foundation model $\theta$, Fine-tuned models $\left\{\theta'_t\right\}_{t=1}^n$, $\alpha$, $\beta$, and $\gamma$}
\KwResult{Multi-task model $\theta^*$}
\For{$t \gets 1$ \KwTo $n$ }{
    {\color{blue} \Comment{Create task direction.}}

    $\theta^d_t \gets \theta'_t - \theta$     
    
    \For{$ \mathrm{layer} \in \mathrm{Layers}(\theta)$  }{
        {\color{blue} \Comment{Record the absolute value of the task direction at the current layer}}

        $p \gets |\theta^d_{t,\mathrm{layer}}| $

        {\color{blue} \Comment{Generate a mask for top k percent of the weights}}
        
        $m_{t,\mathrm{layer}}^\gamma \gets \mathrm{mask\_topk\_percent}(p, k=\gamma)$

        {\color{blue} \Comment{Generate a mask for the bottom k percent of the weights}}

        $m_{t,\mathrm{layer}}^\beta \gets \mathrm{mask\_bottomk\_percent}(p, k=\beta)$

        $m^{\beta, \gamma}_{t,\mathrm{layer}} \gets \mathrm{merge\_masks}(m_{t,\mathrm{layer}}^\beta, m_{t,\mathrm{layer}}^\gamma)$
    }
    {\color{blue} \Comment{Generate 1 mask per fine-tuned model}}
    
    $m^{\beta, \gamma}_t \gets \mathrm{stack\_masks\left(\left\{m^{\beta, \gamma}_{t,\mathrm{layer}}\right\}_{\mathrm{layer} \in \mathrm{Layers}(\theta)}\right)}$

}
{\color{blue} \Comment{Generate the multi-task model}}

$\theta^* \gets \theta + \alpha \sum_{t \in T} m^{\beta, \gamma}_t .  \theta^d_t$

\Return $\theta^*$
\end{algorithm}

\section{Experiments}
\label{sec:exp}
In this section, we conduct a series of experiments to evaluate the Model Breadcrumbs framework. Our experiments focus on the following key aspects:
\begin{enumerate*}
    \item \textbf{Merging Model Breadcrumbs}: We incrementally add tasks, totalling 8 in our investigation, to assess the scalability and performance of merged Model Breadcrumbs as the number of tasks increases.
    \item \textbf{Generalization of Hyperparameters}:  We explore how the hyperparameters introduced by Model Breadcrumbs—$\alpha$, $\beta$, and $\gamma$—generalize over the number of datasets.
    \item \textbf{Effect of Scale}: We investigate the impact of the scale and complexity of the foundation models on the Model Breadcrumbs' adaptability and robustness.
    \item \textbf{Target Task Improvement}: We examine the potential of enhancing the performance a fine-tuned model on a target task by merging related tasks into it.
    \item \textbf{Ablation Study}: We study the importance of the design choices introduced by Model Breadcrumbs for successful and competitive model merging.
\end{enumerate*}

\subsection{Data, Metrics, and Models}
\label{sec:data-and-metrics}
In our analysis, we follow the benchmarks and settings used by Ilharco \etal~\cite{ilharco2022editing} for a more meaningful comparison with existing reports. We present results using normalized accuracy, a metric calculated as the ratio of accuracy achieved by the merged model to that of the fine-tuned model.

\begin{equation}
    \text{Normalized Accuracy} = \frac{\text{Accuracy of Merged Model}}{\text{Accuracy of Fine-tuned Model}}
\end{equation} 

It is noteworthy that the fine-tuned model establishes the upper bound with a normalized accuracy value of 1. Subsequently, the concept of average normalized accuracy is introduced, representing the mean normalized accuracy across multiple tasks. 
In Section~\ref{sec:merging-breadcrumbs}, \ref{sec:validation-free}, \ref{sec:effect-scale}, and \ref{sec:ablation}, we assess our findings using an extensive set of 8 datasets: Cars~\cite{krause20133d}, DTD~\cite{cimpoi14describing}, EuroSAT~\cite{helber2019eurosat}, GTSRB~\cite{Houben-IJCNN-2013}, MNIST~\cite{lecun2010mnist}, RESISC45~\cite{Cheng_2017}, SUN397~\cite{Xiao:2010}, and SVHN~\cite{Netzer2011}. We fine-tune various CLIP models~\cite{radford2021learning} (ViT-B-32, ViT-B-16, and ViT-L-14) to explore model merging. For more information on the datasets and the fine-tuning process see Appendix~\ref{app:vision-data-train}.

In Section~\ref{sec:nlp}, we apply our approach to the NLP domain, specifically investigating four GLUE tasks~\cite{wang-etal-2018-glue} (MRPC~\cite{dolan2005automatically}, RTE~\cite{wang-etal-2018-glue}, CoLA~\cite{warstadt2019neural}, and SST-2~\cite{socher2013recursive}) based on the benchmarks used by \cite{ilharco2022editing, wortsman2022model}. Our process involves fine-tuning the T5-base model~\cite{raffel2020exploring} on these datasets and subsequently merging publicly available fine-tuned models from other datasets (IMDB~\cite{maas2011learning}, RACE~\cite{lai2017race}, QASC~\cite{khot2020qasc}, MultiNews~\cite{fabbri2019multi}, SQuAD~\cite{rajpurkar2016squad}, and CommonGen~\cite{lin2019commongen}) into each of them. Appendix~\ref{app:nlp-data-train} provides additional details on the datasets, the fine-tuning process, and the publicly available fine-tuned models we used in our experiments.

\subsection{Merging Model Breadcrumbs}
\label{sec:merging-breadcrumbs}

\begin{figure}[tb]
\centering
\begin{subfigure}{.49\textwidth}
  \centering
  \includegraphics[width=\linewidth]{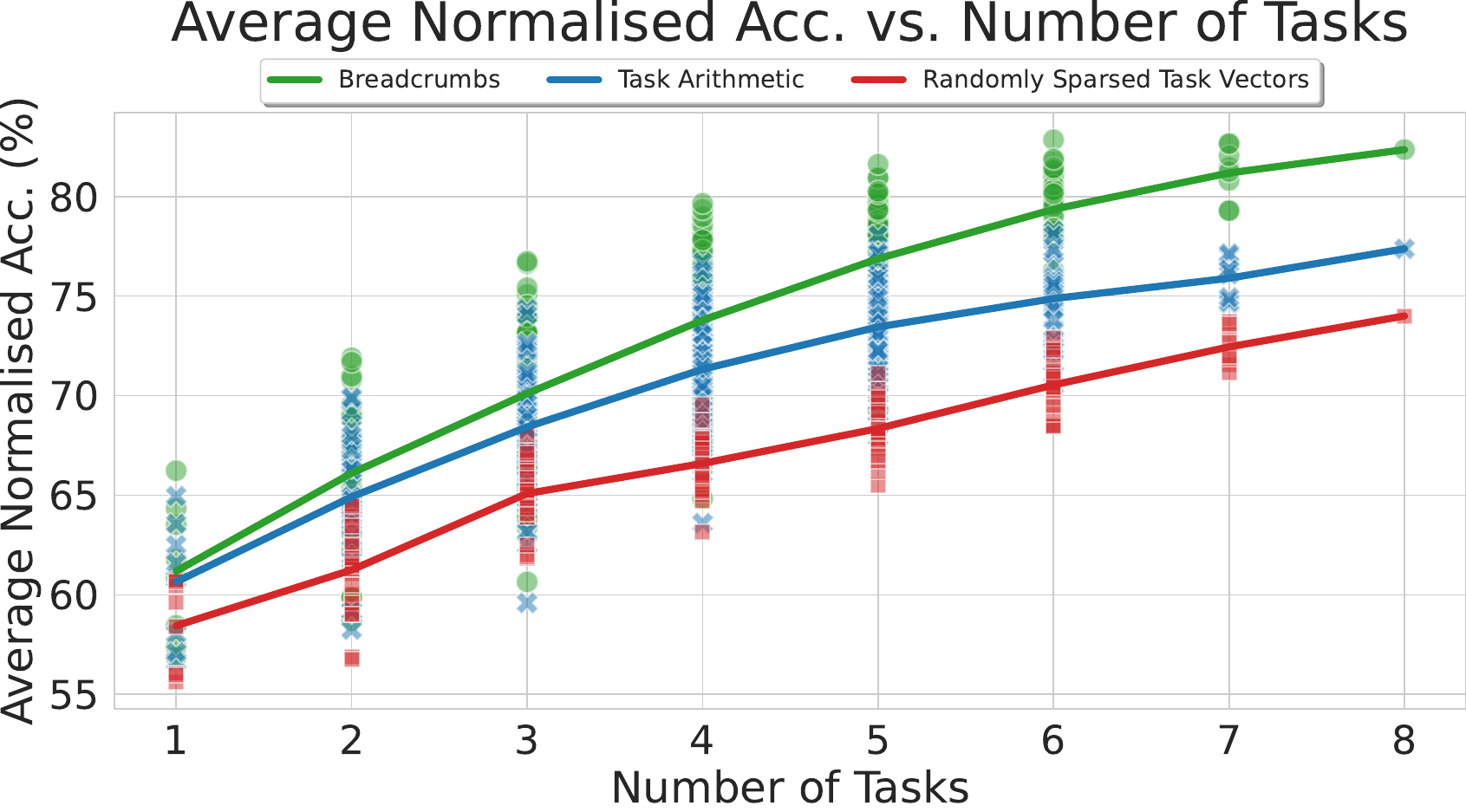}
  \caption{At each point, evaluation is performed over all 8 tasks.}
  \label{fig:vit-b-32-full-and-partial-evaluation-a}
\end{subfigure}\hfill%
\begin{subfigure}{.49\textwidth}
  \centering
  \includegraphics[width=\linewidth]{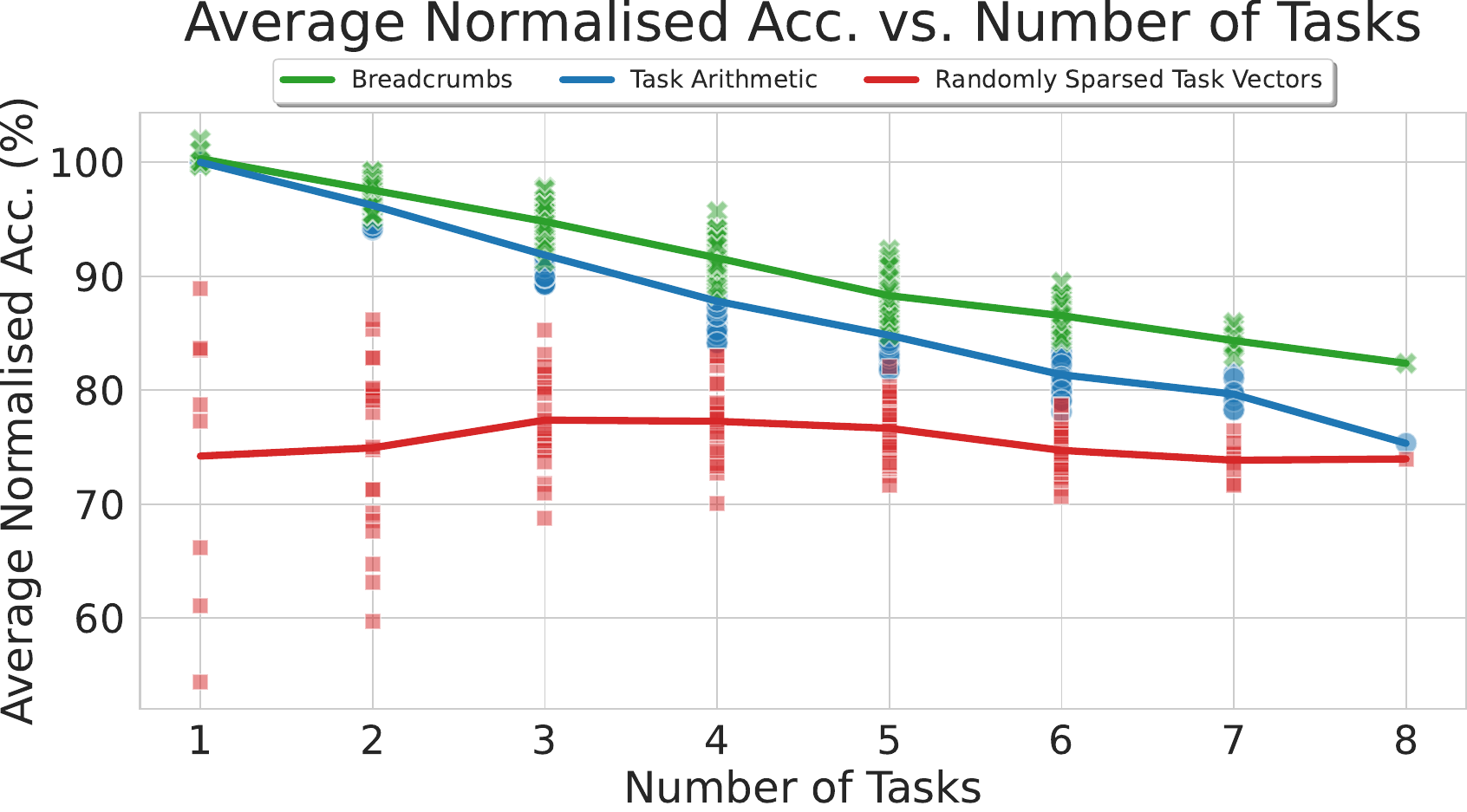}
  \caption{{At each point, evaluation is performed only over the observed tasks.}}
  \label{fig:vit-b-32-full-and-partial-evaluation-b}
\end{subfigure}
  \caption{The solid line is the averaged normalized accuracy across all evaluation points. Each data point corresponds to an experiment involving a subset of the 8 tasks under study. Notably, it is evident that the Model Breadcrumbs (with 90\% sparsity), consistently outperform the Task Arithmetic~\cite{ilharco2022editing}. Specifically, in the experiment involving all eight tasks, the Model Breadcrumbs outperform the Task Arithmetic by a substantial margin of 5.7\%.}
  \label{fig:vit-b-32-full-and-partial-evaluation}
\end{figure}

\begin{figure}[tb]
\centering
\begin{subfigure}{.49\textwidth}
  \centering
  \includegraphics[width=\linewidth]{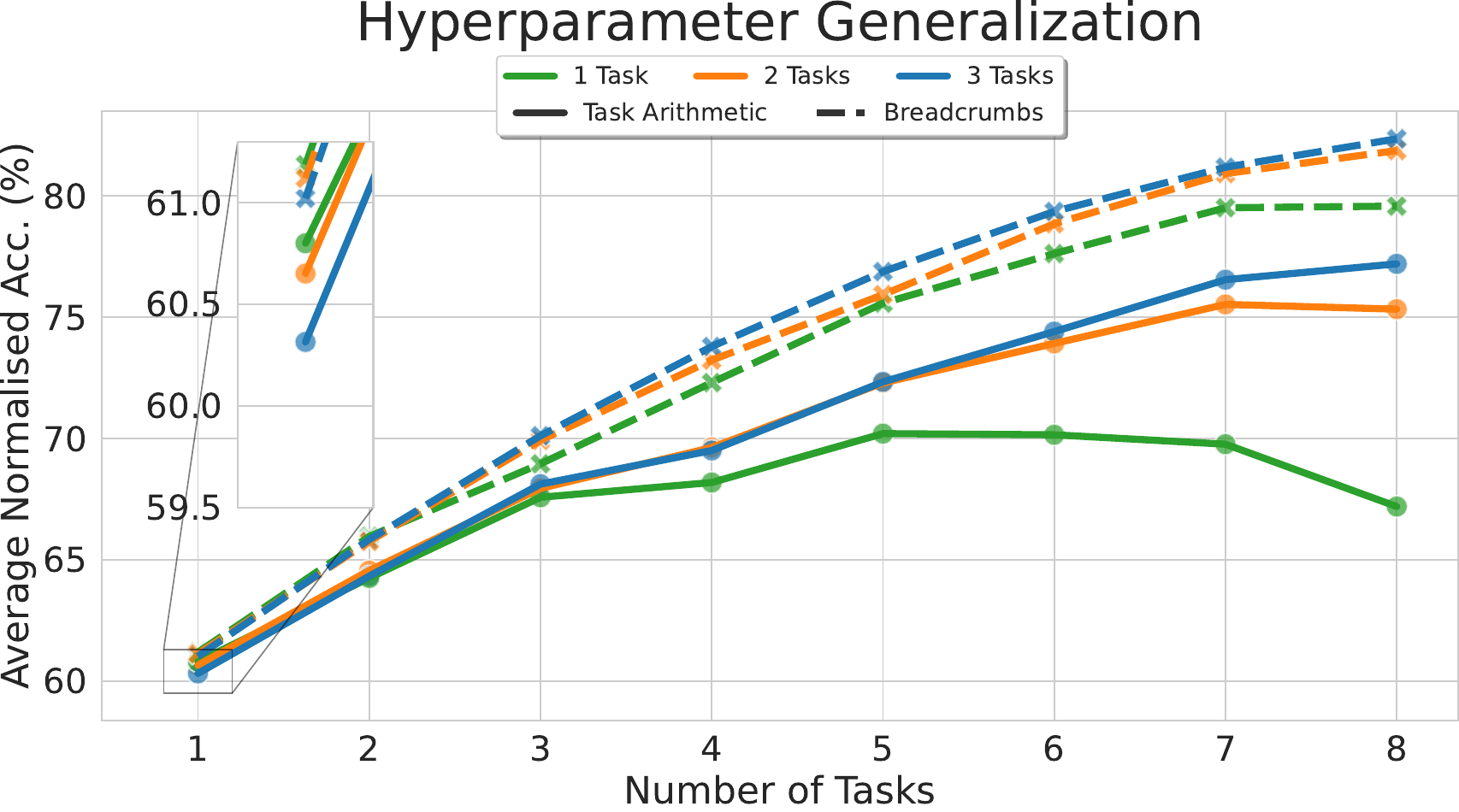}
  \caption{ViT-B-32 Model}
  \label{fig:hparam-b}
\end{subfigure}\hfill%
\begin{subfigure}{.49\textwidth}
  \centering
  \includegraphics[width=\linewidth]{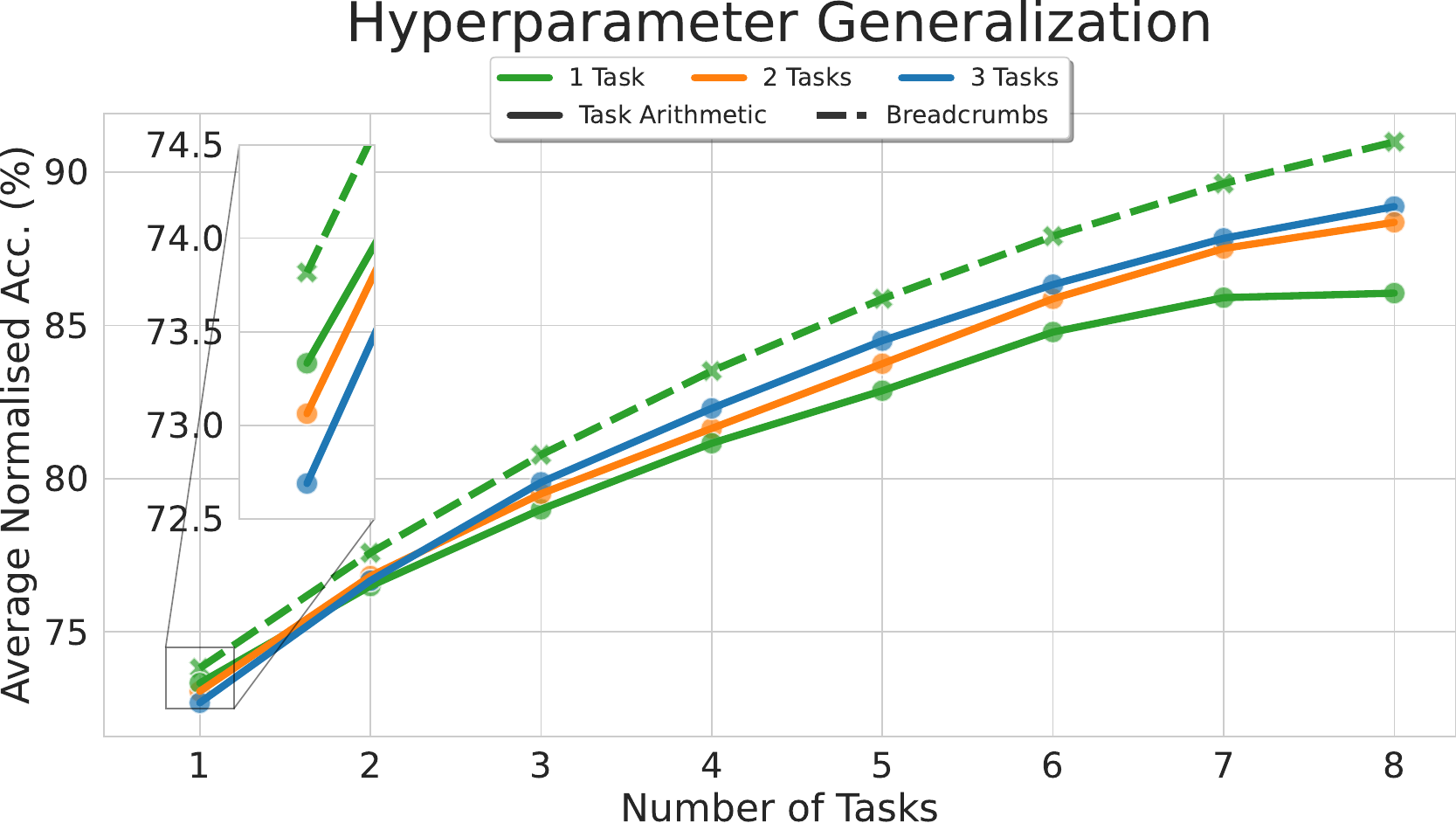}
  \caption{ViT-L-14 Model}
  \label{fig:hparam-l}
\end{subfigure}
  \caption{Validation Free Setting. For the ViT-B-32 model, we tune the hyperparameters of each method (Breadcrumbs and Task Arithmetic) based on the first 1, 2, or 3 tasks and add additional tasks using those hyperparameters (validation set free). For the ViT-L-14 model, the Breadcrumbs method was only tune for the 1 task scenario and evaluate on the additional tasks using those hyperparameters, though the Task Arithmetic appraoch was given more chances to adjust its hyperparameters (task 1, 2, and 3). We observe that Breadcrumbs substantially outperforms task vectors in this setting.}

  \label{fig:hparam-gen}
\end{figure}

\begin{table}[tb]
\caption{The evaluation of the above merging strategies over 8 tasks using ViT-B-32 reveals the advantage of Breadcrumbs over other merging methods. Note that not only Fisher Merging~\cite{NEURIPS2022_70c26937} lags behind both Task Arithmetic~\cite{ilharco2022editing} and Model Breadcrumbs, it also requires significantly more computational resources.}
\label{tab:comp}
\centering
\begin{tabular}{lr}
\toprule
Method    & Avg. Normalized Acc. \\
\midrule
Breadcrumbs                 & \textbf{83.35}               \\
Task Arithmetic~\cite{ilharco2022editing}                & 77.66               \\
Fisher Merging~\cite{NEURIPS2022_70c26937}         & 75.11               \\
Random Sparsed Task Arithmetic & 74.00\\
\bottomrule
\end{tabular}

\end{table}

In this section, we explore the scalability and performance of merged Model Breadcrumbs as we progressively include tasks, reaching a total of 8 in our investigation, as detailed in Section~\ref{sec:data-and-metrics}. Merging enables the creation of multi-task models that can excel across various tasks simultaneously. This versatility is valuable both in scenarios where we have multiple privately fine-tuned models as well as in cases where we have access to publicly available fine-tuned models. This allows the extraction of existing knowledge from these models without the need for extra training or access to additional training data. Table~\ref{tab:comp} presents a comparison between Model Breadcrumbs with 90\% sparsity ($\beta=90\%$, $\gamma=99\%$), the recently proposed Task Arithmetic~\cite{ilharco2022editing}, and Fisher Merging~\cite{NEURIPS2022_70c26937} across 8 tasks, using ViT-B-32 model. Model Breadcrumbs outperforms all considered methods by a substantial margin. Fisher Merging~\cite{NEURIPS2022_70c26937} lags behind both Task Arithmetic~\cite{ilharco2022editing} and Model Breadcrumbs, and notably, it requires significantly more computational resources. Therefore, we proceed with the rest of our studies without evaluating Fisher Merging.

In Figure~\ref{fig:vit-b-32-full-and-partial-evaluation}, we assess all possible task subsets of the 8 tasks detailed in Section~\ref{sec:data-and-metrics}, amounting to a total of $256=2^8$ combinations, under two settings:
\begin{enumerate*}
    \item evaluation over all 8 tasks and,
    \item evaluation only on the subset of tasks that have been observed.
\end{enumerate*}
As we can see in Figure \ref{fig:vit-b-32-full-and-partial-evaluation-a} merging Model Breadcrumbs (90\% sparsity) results in superior multi-task models compared to the Task Arithmetic~\cite{ilharco2022editing}. Furthermore, the performance gap between these two approaches increases as more tasks are observed, resulting in vastly superior multi-task models when more Model Breadcrumbs are available.

In Figure~\ref{fig:vit-b-32-full-and-partial-evaluation-b} we can see that for small task numbers the resulting merged model performs closely to that of the multiple fine-tuned models although the gap increases as more tasks are added. Model Breadcrumbs again prove to be more performance that Task Arithmetic~\cite{ilharco2022editing} in this setting.

\subsection{Validation-Free Setting}
\label{sec:validation-free}

\begin{figure}[tb]
    \centering
    \includegraphics[width=\linewidth]{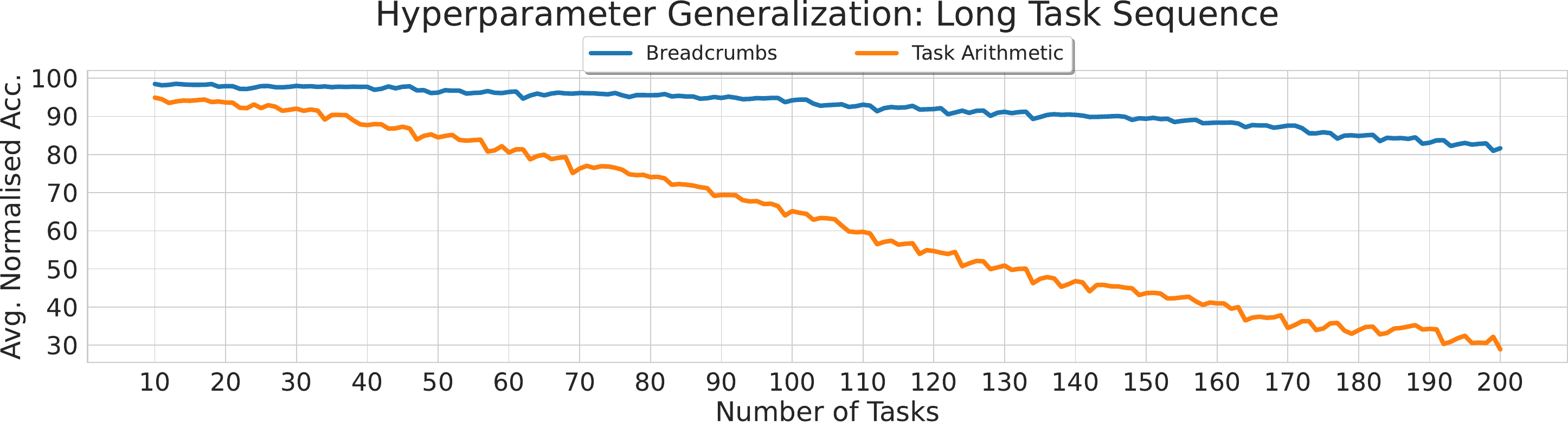}

    \caption{The 200-task sequence originates from the ImageNet dataset~\cite{deng2009imagenet}, created by dividing the data into 200 5-class classification tasks. After encountering 10 tasks using the ViT-L-14 model, the best hyperparameters for each method (Breadcrumbs with 85\% sparsity and Task Arithmetic~\cite{ilharco2022editing}) are selected and fixed. Each point on the plot represents the evaluation of the method over all tasks observed up to that point. With an increasing number of tasks, Model Breadcrumbs consistently outperforms Task Arithmetic~\cite{ilharco2022editing} by a substantial margin, highlighting the robustness of hyperparameters in the Model Breadcrumbs approach.}

    \label{fig:split-imagenet}
\end{figure}

\begin{figure}[tb]
\centering
\begin{subfigure}{.49\textwidth}
  \centering
  \includegraphics[width=\linewidth]{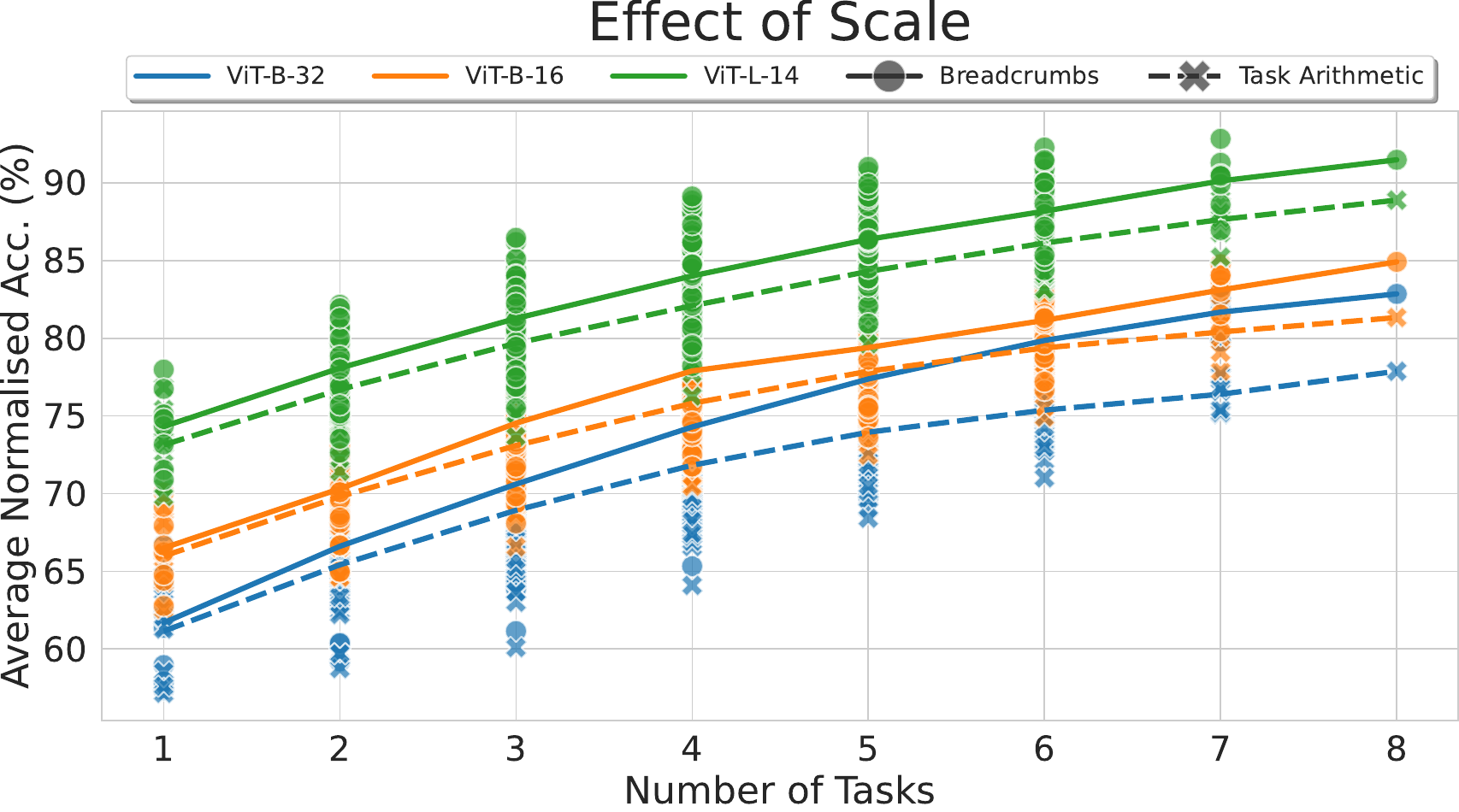}
  \caption{At each point, evaluation is performed over all 8 tasks.}
  \label{fig:scale-full}
\end{subfigure}\hfill%
\begin{subfigure}{.49\textwidth}
  \centering
  \includegraphics[width=\linewidth]{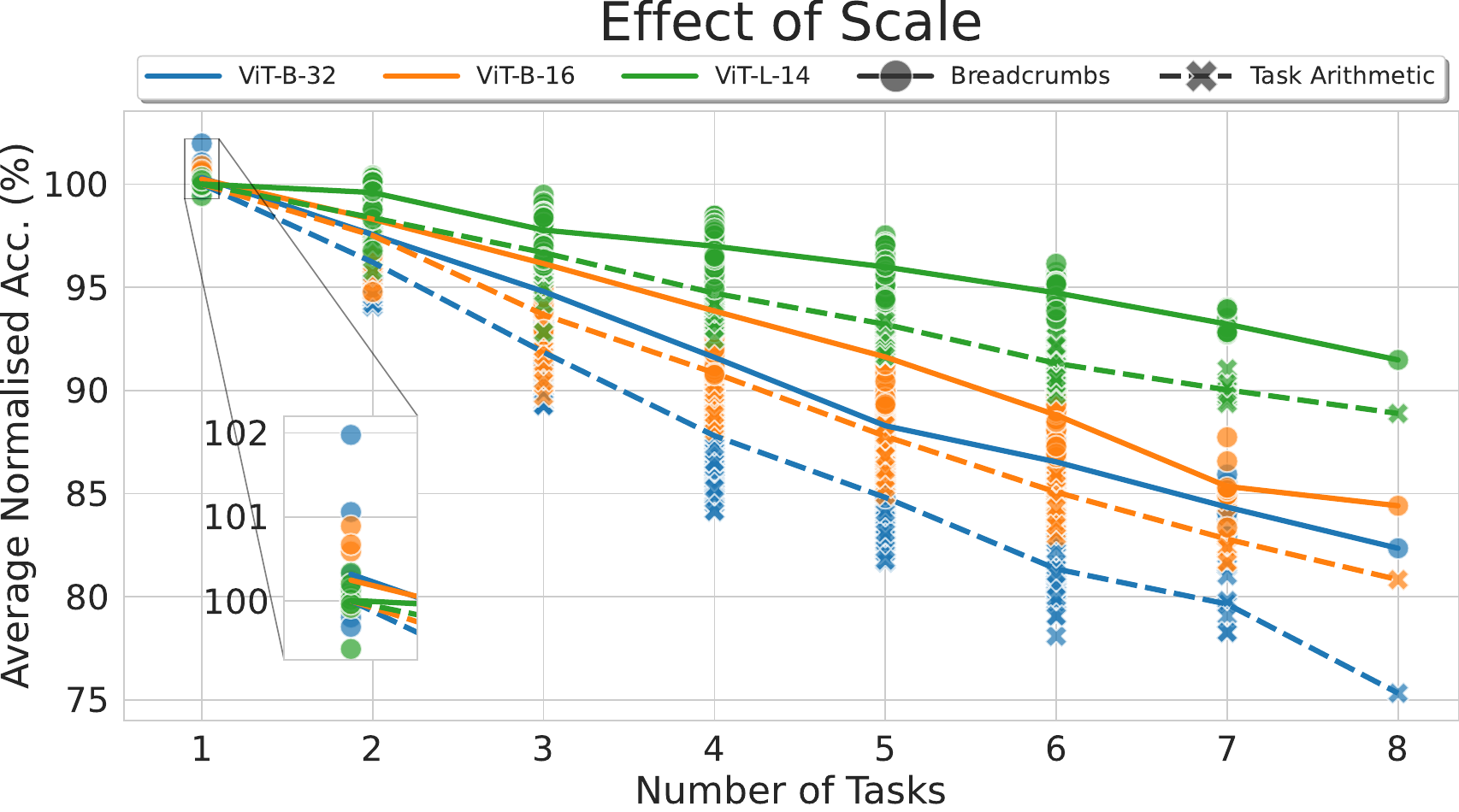}
  \caption{At each point, evaluation is performed only over the observed tasks.}
  \label{fig:scale-partial}
\end{subfigure}

  \caption{Comparative performance analysis of Model Breadcrumbs and Task Arithmetic~\cite{ilharco2022editing} methods across varying CLIP model scales (ViT-B-32, ViT-B-16, and ViT-L-14) as the number of tasks increases. The solid line represents the averaged normalized accuracy across all evaluation points. Each data point corresponds to an experiment involving a subset of the 8 tasks under study. Our findings highlight the potential of larger-scale models to mitigate performance degradation and, as seen in Figure~\ref{fig:scale-partial}, the capability of Model Breadcrumbs to produce multi-task models that surpass individual fine-tuned models for specific tasks.}

  \label{fig:scale}
\end{figure}

In Section~\ref{sec:merging-breadcrumbs}, we compared Model Breadcrumbs and Task Arithmetic~\cite{ilharco2022editing} under their respective optimal hyperparameters. These hyperparameters were fine-tuned based on model performance on the validation dataset for each subset of tasks following \cite{ilharco2022editing}. However, as the number of tasks increases, the search for optimal hyperparameters becomes increasingly resource-intensive. Furthermore, the need for a validation set from each task being added can be restrictive due to privacy concerns or due to the unavailability of additional validation data. Thus, we consider a new setting where hyperparamters are tuned based on a few tasks, and subsequent tasks are added using these pre-determined hyperparameters.

The results are shown in Figure~\ref{fig:hparam-gen}. Remarkably, our experiments reveal that the hyperparameters of Model Breadcrumbs exhibit a high degree of generalizability. Specifically, for the ViT-B-32 model when considering scenarios involving three tasks and beyond, up to the 8-task scenario, the optimal hyperparameters remain consistent. Moreover, for the ViT-L-14 model, the hyperparameters do not change beyond the 1 task scenario. This remarkable stability underscores the robustness and versatility of Model Breadcrumbs. We observer that on the other hand the approach Task Arithmetic~\cite{ilharco2022editing} can quickly collapse in performance.

Motivated by these results, we extended the evaluation to a much longer task sequence using ViT-L-14 model. We split the ImageNet data~\cite{deng2009imagenet} into 200 tasks, each classifying 5 classes. After finding optimal hyperparameters for both Model Breadcrumbs and Task Arithmetic using 10 tasks, we kept these hyperparameters and incrementally merged all 200 tasks to create a multi-task model. As seen in Figure~\ref{fig:split-imagenet}, the observed trend remains, with Model Breadcrumbs (85\% sparsity: $\beta=85\%$, $\gamma=99.3\%$) consistently outperforming Task Arithmetic~\cite{ilharco2022editing} by a significant margin as the number of tasks increases. This showcases the generalizability of the hyperparameters for the Model Breadcrumbs approach.

The practical implication of this stability in hyperparameter settings is that, in practice, we can rely on a relatively small number of tasks to determine optimal hyperparameters when applying Model Breadcrumbs to diverse multi-task learning scenarios. This simplifies the implementation process, reduces the need for extensive hyperparameter tuning, and contributes to the framework's practicality and ease of use.
In contrast, Task Arithmetic~\cite{ilharco2022editing} do not exhibit the same level of hyperparameter stability. Consequently, this fundamental divergence between Model Breadcrumbs and Task Arithmetic~\cite{ilharco2022editing} underlines the substantial advantage of Model Breadcrumbs in real-world multi-task learning scenarios.

\subsection{Effect of Scale}
\label{sec:effect-scale}
\begin{figure}[tb]
    \centering
    \includegraphics[width=\linewidth]{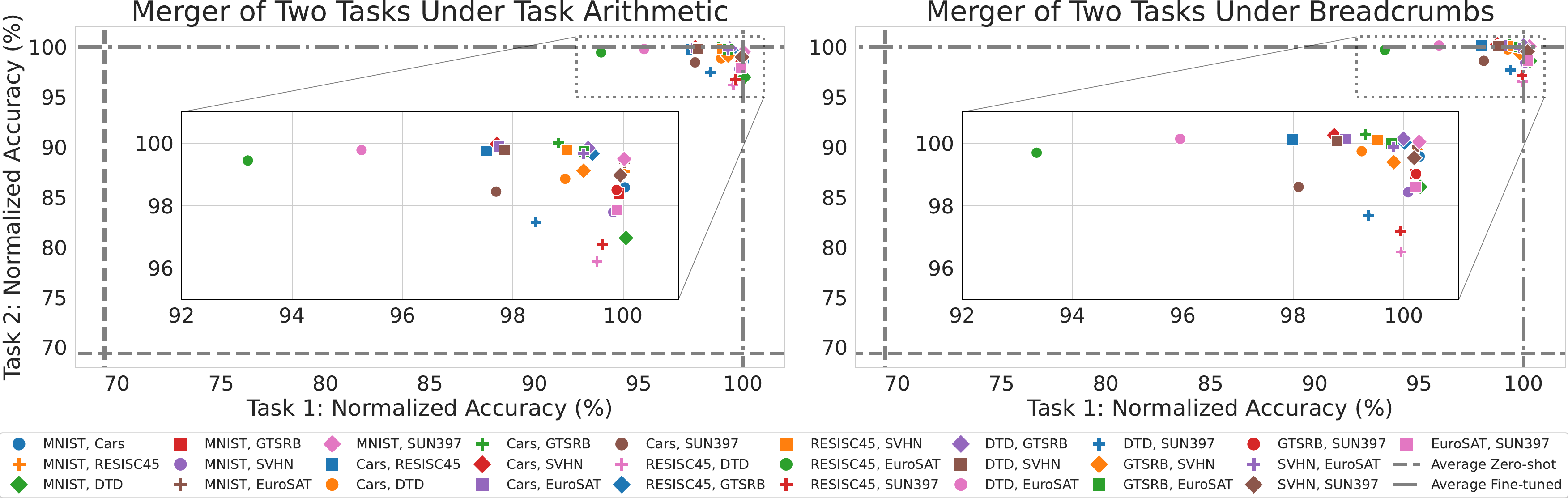}

    \caption{Comparison of Model Breadcrumbs and Task Arithmetic~\cite{ilharco2022editing} in the merger of task pairs, revealing improved accuracy on both tasks and a higher frequency of multi-task models surpassing individual fine-tuned accuracy levels when employing Model Breadcrumbs.}

    \label{fig:2-tasks}
\end{figure}

In this section, we explores the impact of using larger CLIP models on our analysis, comparing the performance of ViT-B-32, ViT-B-16, and ViT-L-14 models. For each model type, the optimal Model Breadcrumbs were found at 90\% ($\beta=90\%$, $\gamma=99\%$), 90\% ($\beta=90\%$, $\gamma=99.2\%$), 85\% ($\beta=85\%$, $\gamma=99\%$) sparsity respectively. As shown in Figure~\ref{fig:scale}, the adoption of larger models significantly improves the performance of both our proposed Model Breadcrumbs method and the Task Arithmetic~\cite{ilharco2022editing} baseline. Moreover, as more tasks are introduced, the capacity to construct better-performing multi-task models grows, with larger-scale models demonstrating superior results.

Specifically, we observe in Figure~\ref{fig:scale-full}, when utilizing the ViT-L-14 model and considering 8 tasks, merging Model Breadcrumbs produces a single multi-task model with an average performance that reaches 91.48\% of the performance achieved by employing 8 individual fine-tuned models (i.e., one per task). The shift from 8 fine-tuned models to a single multi-task model substantially reduces inference time and compute resources, accompanied by only a minor relative loss in performance. This underscores the practical advantages of our approach.

Moreover, Figure~\ref{fig:scale-partial} highlights that the performance decline observed when merging either Model Breadcrumbs or Task Arithmetic~\cite{ilharco2022editing} can be significantly mitigated by adopting larger-scale models. Notably, for the ViT-L-14 model, merging Model Breadcrumbs for certain tasks can result in multi-task models that either match or surpass the performance of individual fine-tuned models. To delve deeper into this phenomenon, we conducted a closer examination of task merger for ViT-L-14, considering the two tasks scenario.

As we can see in Figure~\ref{fig:2-tasks}, when adding pairs of tasks via Model Breadcrumbs and Task Arithmetic~\cite{ilharco2022editing}, the merger generally leads to improved performance on both tasks, resulting in a single model that is competitive and often superior to using two specialized fine-tuned models. Furthermore, for the same task pairs, Model Breadcrumbs consistently produces multi-task models that surpass their equivalent Task Arithmetic~\cite{ilharco2022editing} versions. Notably, Model Breadcrumbs mergers generate a higher number of multi-task models where both tasks exceeded their respective fine-tuned accuracy levels. This highlights the potential of Model Breadcrumbs not only to maintain but also to enhance task-specific performance within a multi-task framework. We further examine this concept in the next section.

\subsection{Target Task Improvement via Model Merging}
\label{sec:nlp}
\begin{table}[tb]
\caption{Model merging enhances the fine-tuned models. Specifically, the merger of Breadcrumbs yields higher-performing models without requiring additional training data or combining with models trained on similar data. Values represent the average performance over 20 runs, followed by the standard error.}

\label{tab:nlp}
\centering
\begin{tabular}{l@{\hskip 1em}c@{\hskip 1em}c@{\hskip 1em}c@{\hskip 1em}r}
\toprule
Method / Dataset          & MRPC & RTE  & CoLA & SST-2 \\
\midrule
Zero-shot                & 74.8 & 52.7 & 8.29 & 92.7  \\
Fine-tuned           & 87.9 $\pm 0.68$  & 76.2 $\pm 0.46$& 51.6 $\pm 0.37$& 93.3 $\pm 0.35$ \\
Fine-tuned + Task Arithmetic & 88.6 $\pm 0.59$ & 76.4 $\pm 0.40$& 52.1 $\pm 0.32$& 93.5 $\pm 0.31$\\
Fine-tuned + Breadcrumbs  & \textbf{90.0} $\mathbf{\pm 0.50}$ & \textbf{77.5} $\mathbf{\pm 0.38}$& \textbf{53.2} $\mathbf{\pm 0.34}$& \textbf{94.4} $\mathbf{\pm 0.32}$\\
\bottomrule
\end{tabular}

\end{table}

Motivated by the insights from Figure~\ref{fig:2-tasks}, we explore the potential of enhancing the performance of a fine-tuned model for a specific target task solely through model merging. We fine-tune the T5-base model~\cite{raffel2020exploring} for four GLUE tasks~\cite{wang-etal-2018-glue} (discussed in Section~\ref{sec:data-and-metrics}) based on benchmarks used by \cite{ilharco2022editing, wortsman2022model}. We then merge six publicly available T5-base models (IMDB~\cite{maas2011learning}, RACE~\cite{lai2017race}, QASC~\cite{khot2020qasc}, MultiNews~\cite{fabbri2019multi}, SQuAD~\cite{rajpurkar2016squad}, and CommonGen~\cite{lin2019commongen}) with each of them. Appendix~\ref{app:nlp-data-train} provides more details on fine-tuned models and the fine-tuning process.

The results, presented in Table~\ref{tab:nlp}, evaluates various approaches, including Zeroshot, pure fine-tuning, utilizing Task Arithmetic~\cite{ilharco2022editing} derived from newly added tasks in the fine-tuned models, and employing Breadcrumbs of the new tasks added to the fine-tuned models. The findings demonstrate that incorporating Breadcrumbs from these new tasks effectively enhances the performance of our fine-tuned models, surpassing all other considered approaches. Importantly, this improvement is achieved without the need for additional training or requiring data from the exact same dataset. This approach underscores the versatility and effectiveness of utilizing Model Breadcrumbs to improve task performance across diverse tasks.

\subsection{Ablations}
\label{sec:ablation}

\begin{figure}[tb]
  \begin{minipage}[c]{0.55\textwidth}
  \centering
    \includegraphics[width=\textwidth]{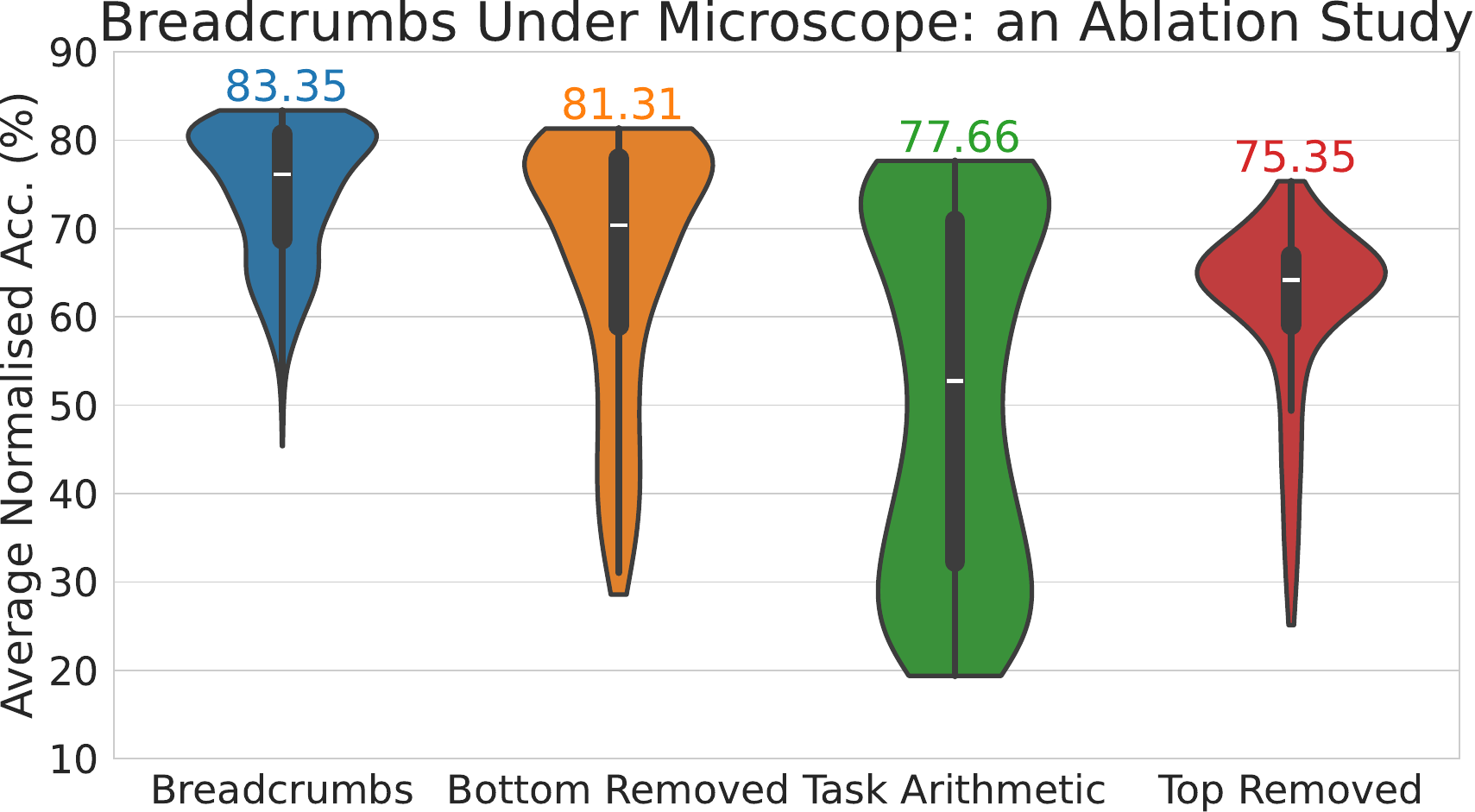}
  \end{minipage}
  \begin{minipage}[c]{0.44\textwidth}
    \caption{Performance comparison of the Model Breadcrumbs against alternative masking choices, reveals: Model Breadcrumbs yields a higher distribution of high-performance multi-task models, underlining its robustness towards hyperparameter perturbations. Model Breadcrumbs produces the highest performing multi-task model. The number on top of each violin indicates the performance of the highest performing model of that setting.
    } 
    \label{fig:ablation}
  \end{minipage}
\end{figure}

\begin{figure}[tb]

    \centering
    \includegraphics[width=0.9\linewidth]{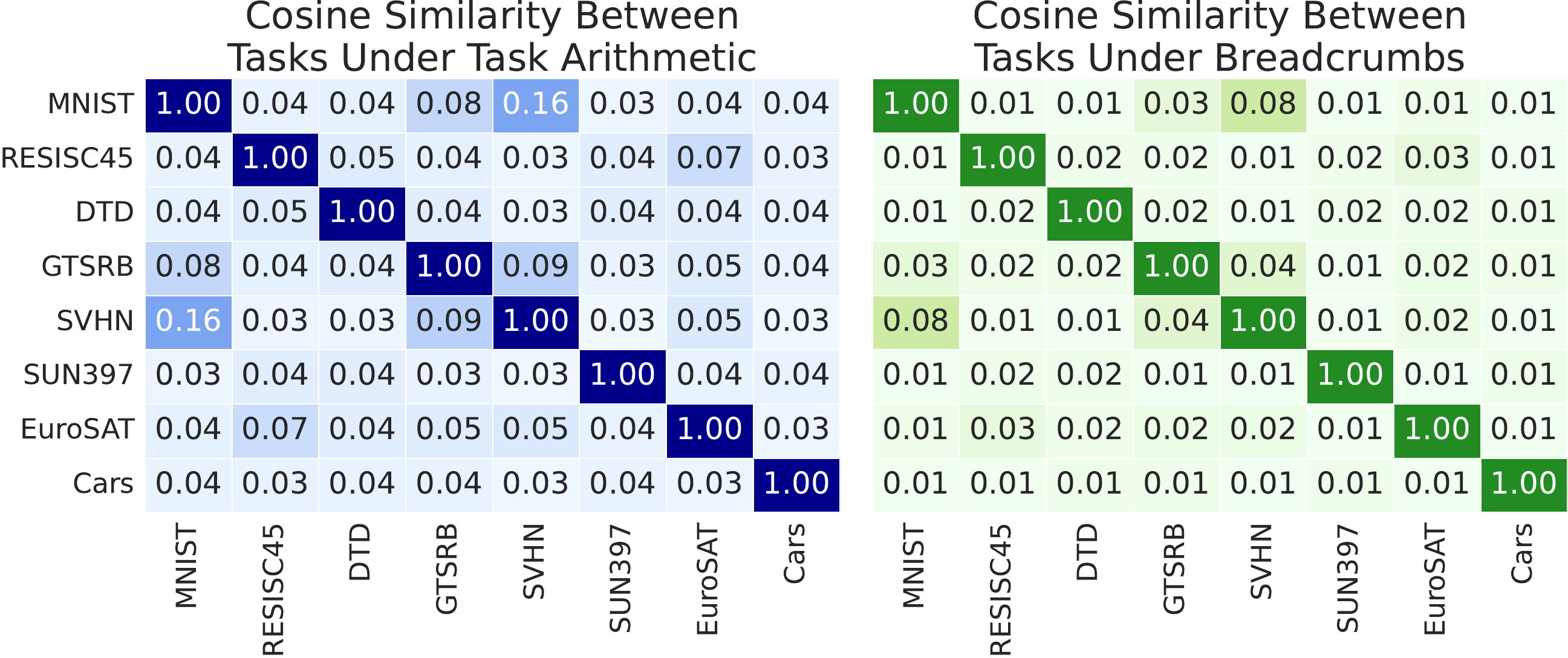}
    \caption{Comparison of Cosine Similarity Between Tasks in Model Breadcrumbs and Task Arithmetic. The figure illustrates the cosine similarity distribution among tasks, highlighting how Model Breadcrumbs enforces greater orthogonality, leading to reduced interference during model merging.}
    \label{fig:similarity}

\end{figure}

In this section, we perform ablations to examine alternative design decisions within the Model Breadcrumbs method. Specifically, we explore different approaches for constructing the masking operation, namely:
\begin{enumerate*}
    \item Bottom-Weight Masking: Masking only the bottom-most smallest absolute magnitude weights per layer.
    \item Top-Weight Masking: Masking only the top largest absolute magnitude weights per layer.
\end{enumerate*}
We compare these alternatives to the full Model Breadcrumbs approach, which encompasses both (1) and (2), as well as the Task Arithmetic~\cite{ilharco2022editing} method, which lacks any masking. In our investigation, we conduct a grid search to identify the optimal hyperparameters for each of the four configurations. We assess the resulting multi-task models on 8 tasks discussed in Section~\ref{sec:data-and-metrics}. The results are shown in Figure~\ref{fig:ablation}.

Our findings reveal two key insights:
\begin{enumerate*} [label=(\roman*)]
    \item both forms of weight masking, as employed in Model Breadcrumbs, are essential for achieving competitive performance. Model Breadcrumbs, which combines both bottom and top weight masking, emerges as the most effective approach.
    \item The grid search for hyperparameters within the Model Breadcrumbs approach yields a higher distribution of high-performance multi-task models compared to the other three settings. Furthermore, there is much lower variation in the overall performance distribution of the multi-task models produced by the Model Breadcrumbs.
    These observations underscore the robustness of Model Breadcrumbs to variations in hyperparameter settings, further enhancing its practicality and reliability in real-world applications.
\end{enumerate*}

In Figure~\ref{fig:similarity}, we examine the cosine similarity between tasks using Model Breadcrumbs and Task Arithmetic~\cite{ilharco2022editing}. Most tasks show orthogonality, indicating minimal side effects upon merging. However, upon closer inspection, semantically similar tasks (e.g., MNIST~\cite{lecun2010mnist}, SVHN~\cite{Netzer2011}, and GTSRB~\cite{Houben-IJCNN-2013}) exhibit higher cosine similarity, suggesting non-orthogonality. This similarity could introduce interference during merging. In contrast, Model Breadcrumbs pushes all cosine similarity values closer to zero, reinforcing orthogonality. This reduction in interference could explain the enhance performance of the resulting multi-task models when using Breadcrumbs.

\begin{figure}[tb]
\centering
\begin{subfigure}{.49\textwidth}
  \centering
  \includegraphics[width=\linewidth]{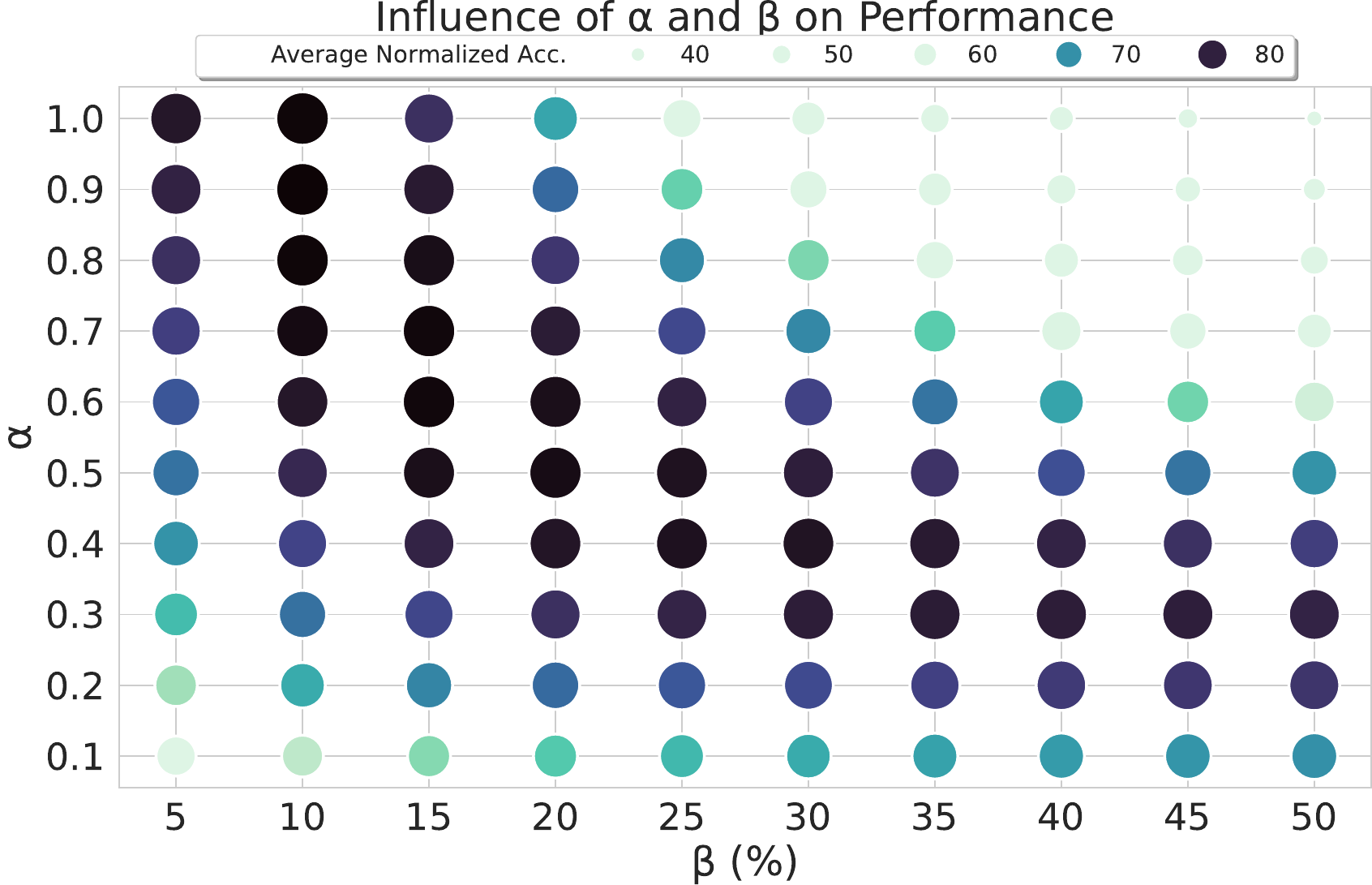}
  \caption{$\alpha$ vs. $\beta$: as $\beta$ grows in magnitude lower $\alpha$'s are required.}
  \label{fig:alpha-beta}
\end{subfigure}\hfill%
\begin{subfigure}{.49\textwidth}
  \centering
  \includegraphics[width=\linewidth]{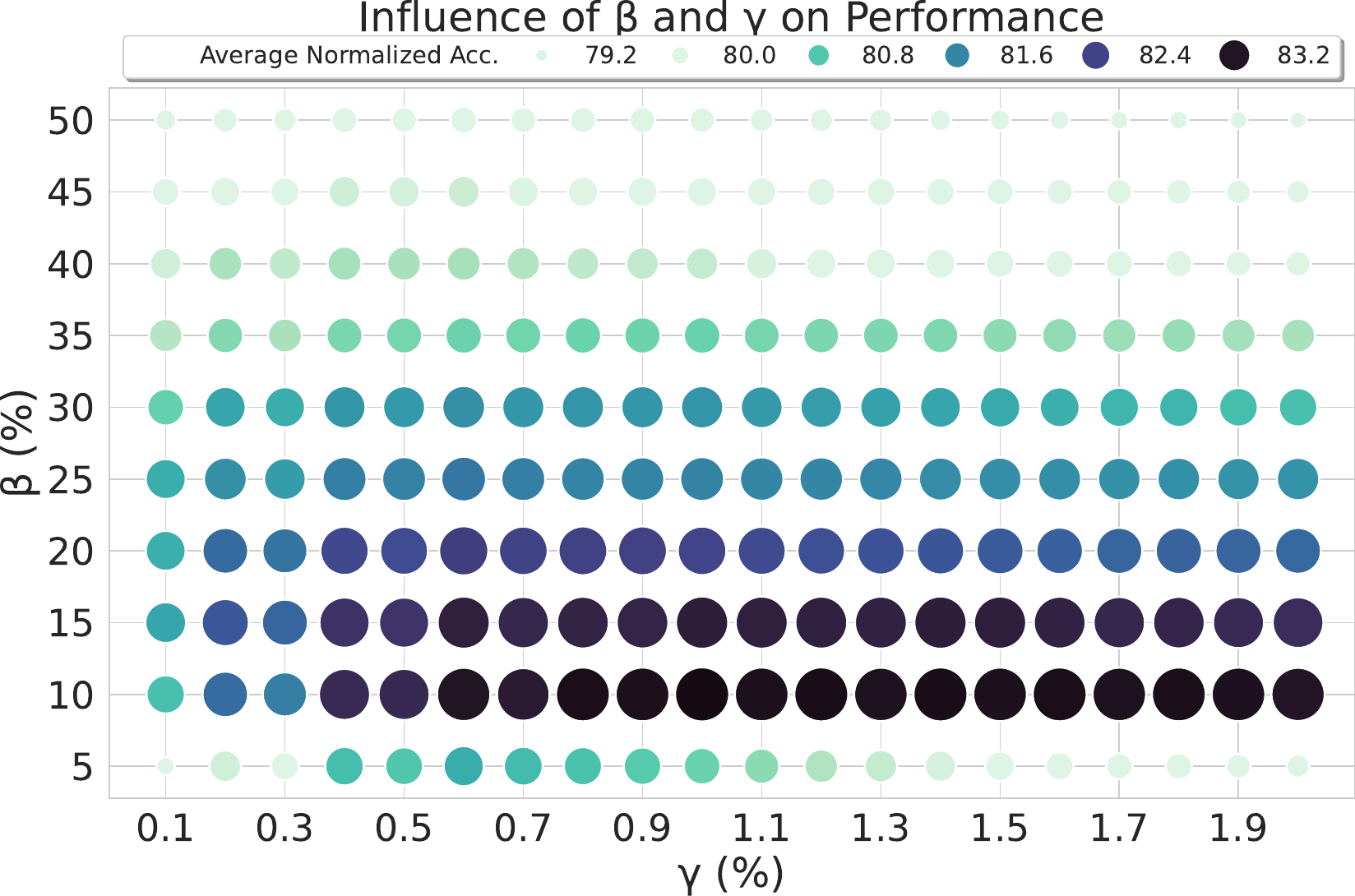}
  \caption{$\beta$ vs. $\gamma$: lower $\beta$'s have more tolerance on variations of $\gamma$.}
  \label{fig:beta-gama}
\end{subfigure}

  \caption{Influence of hyperparameters on model performance using the ViT-B-32 model across eight vision tasks. For ease of readability, in this part we use $\beta$ and $\gamma$ to represent how much weights have been masked. It shows the relationships between $\alpha$ and $\beta$, and between $\beta$ and $\gamma$, highlighting the stability of hyperparameters across the possible combinations.}
    \label{fig:breadcrumbs-hyperparams}

\end{figure}

In Figure~\ref{fig:breadcrumbs-hyperparams}, we demonstrate the impact of hyperparameters on the performance of models using the ViT-B-32 model, assessed across eight vision tasks outlined in Section~\ref{sec:data-and-metrics}. For ease of readability, in this part we use $\beta$ and $\gamma$ to represent how much weights have been masked. Figure~\ref{fig:alpha-beta} examines the relationship between $\alpha$ and $\beta$, the primary determinants of task vector sparsity. As $\beta$ increases and more weights are masked, large alphas, which amplify the remaining weights' contributions, become less tolerable, necessitating lower $\alpha$'s as $\beta$ grows. Upon identifying optimal $\alpha$ and $\beta$ values, we investigate gamma. In Figure~\ref{fig:beta-gama}, we depict the relationship between $\beta$ and gamma. Regardless of beta's value, the $\gamma$ that optimizes a combination of $\alpha$ and $\beta$ tends to hover around 1\%, with lower betas allowing for higher gammas and vice versa. Across both figures, we consistently observe that numerous combinations of alpha, beta, and $\gamma$ result in high-performing merged models, as previously noted in Figure~\ref{fig:ablation}.

\section{Conclusions}
\label{sec:conclusion}
In this paper, we introduced Model Breadcrumbs, a simple yet effective approach to constructing multi-task models from pre-existing fine-tuned foundation models. Our extensive experiments showcase the method's capability to enhance performance across multiple tasks, demonstrating stable and generalizable hyperparameters. This simplicity makes Model Breadcrumbs practical for real-world multi-task learning scenarios. Additionally, scaling experiments indicate that larger models further benefit from the approach, narrowing the performance gap between merged models and individual fine-tuned ones. Notably, our exploration in NLP data highlights the method's versatility across different modalities.

While promising, Model Breadcrumbs has limitations. Its performance hinges on the quality of the initial fine-tuned models; issues like poor generalization or severe overfitting can propagate. Future research can delve into mitigating these limitations and exploring more sophisticated aggregation techniques for multiple trajectories. Additionally, as the number of tasks increases, considering the expansion of model capacity becomes crucial for sustained high performance.

In conclusion, Model Breadcrumbs stands out for its simplicity, efficiency, and effectiveness in constructing multi-task models. Leveraging publicly available fine-tuned models, it aligns with the trend of updatable machine learning, supporting community-driven model refinement efforts. We anticipate that Model Breadcrumbs will contribute to the development of efficient and scalable multi-task learning solutions in the future.

\section*{Acknowledgements}
We acknowledge funding from the NSERC Discovery Grant RGPIN-2021-04104 and FRQNT New Scholar. This research was enabled in part by compute resources provided by Digital Research Alliance of Canada (the Alliance) and Calcul Québec.

\newpage
\bibliographystyle{splncs04}
\bibliography{main}

\newpage

\appendix
\section{Vision: Data and Training}
\label{app:vision-data-train}

We fine-tune CLIP models~\cite{radford2021learning} using the datasets specified in Section~\ref{sec:data-and-metrics} (more details are given in Tabel~\ref{table:1}). Following a procedure akin to \cite{ilharco2022patching}, our fine-tuning comprises 2000 iterations with a batch size of 128, a learning rate set to 1e-5, and a cosine annealing learning rate schedule with 200 warm-up steps. The AdamW optimizer~\cite{loshchilov2017decoupled} with a weight decay of 0.1 is employed for optimization.

Throughout the fine-tuning process, we freeze the weights of CLIP's text encoder classification layer. This ensures no introduction of additional learnable parameters, a strategy validated in prior work~\cite{ilharco2022patching}.

\begin{table}[tb]
\centering
\begin{tabular}{lrrrr}
\toprule
Dataset  & Training & Validation & Testing & Classes \\
\midrule
Cars     & 7,330    & 814        & 8041    & 196               \\
DTD      & 3,384    & 376        & 1,880   & 47                \\
EuroSAT  & 21,600   & 2,700      & 2,700   & 10                \\
GTSRB    & 23,976   & 2,664      & 12,630  & 43                \\
MNIST    & 55,000   & 5,000      & 10,000  & 10                \\
RESISC45 & 17,010   & 1,890      & 6,300   & 45                \\
SUN397   & 17,865   & 1,985      & 19,850  & 397               \\
SVHN     & 68,257   & 5,000      & 26,032  & 10 \\     
\bottomrule
\end{tabular}
\caption{Data statistics.}
\label{table:1}
\end{table}

\section{NLP: Data and Training}
\label{app:nlp-data-train}

We fine-tune the T5-base model~\cite{raffel2020exploring} on MRPC~\cite{dolan2005automatically}, RTE~\cite{wang-etal-2018-glue}, CoLA~\cite{warstadt2019neural}, and SST-2~\cite{socher2013recursive}. Our fine-tuning process utilizes a batch size of 32, a learning rate of 1e-5, and lasts for 5 epochs using the AdamW optimizer with a linear learning rate schedule. We use a create a validation set from the training data equal in size to the test set to pick the best model. The publicly available fine-tuned models are sourced from the Hugging Face hub\footnote{\url{https://huggingface.co/models
}}, and the specific models can be accessed via the following links:
\begin{itemize}
    \item  IMDB: \texttt{mrm8488/t5-base-finetuned-imdb-sentiment}
    \item RACE: \texttt{mrm8488/t5-base-finetuned-race}
    \item QASC: \texttt{mrm8488/t5-base-finetuned-qasc}
    \item MultiNews: \texttt{mrm8488/t5-base-finetuned-summarize-news}
    \item SQuAD: \texttt{mrm8488/t5-base-finetuned-question-generation-ap}
    \item CommonGen: \texttt{mrm8488/t5-base-finetuned-common gen}
\end{itemize}

\section{TIES Merging}
\label{us-vs-ties}
A concurrent study by Yadav \etal~\cite{yadav2023ties} presents a method named TIES. Like the Task Arithmetic method~\cite{ilharco2022editing}, TIES initially constructs a set of Task Vectors. These vectors undergo a masking process to eliminate interfering weights, identified as a percentage of overall weights with low magnitudes. The remaining unmasked weights undergo a sign alignment operation to determine their polarity. Finally, a scaled sum merges the task vectors with the pre-trained model.

Our approach differs from TIES in two key aspects. Firstly, we apply masking to both very large and small magnitude weights of the task vectors to minimize interference, whereas TIES focuses solely on small magnitude weights. Secondly, our masking strategy employs layer-wise masking, wherein a certain percentage of weights are masked based on their magnitude relative to the weights within that layer, as opposed to overall masking, which ranks all model weights by magnitude and masks the smallest ones. Notably, in the context of task vectors, overall masking typically targets weights in the early layers~\cite{NEURIPS2022_70c26937}.

\begin{figure}[tb]

    \centering
    \includegraphics[width=0.8\linewidth]{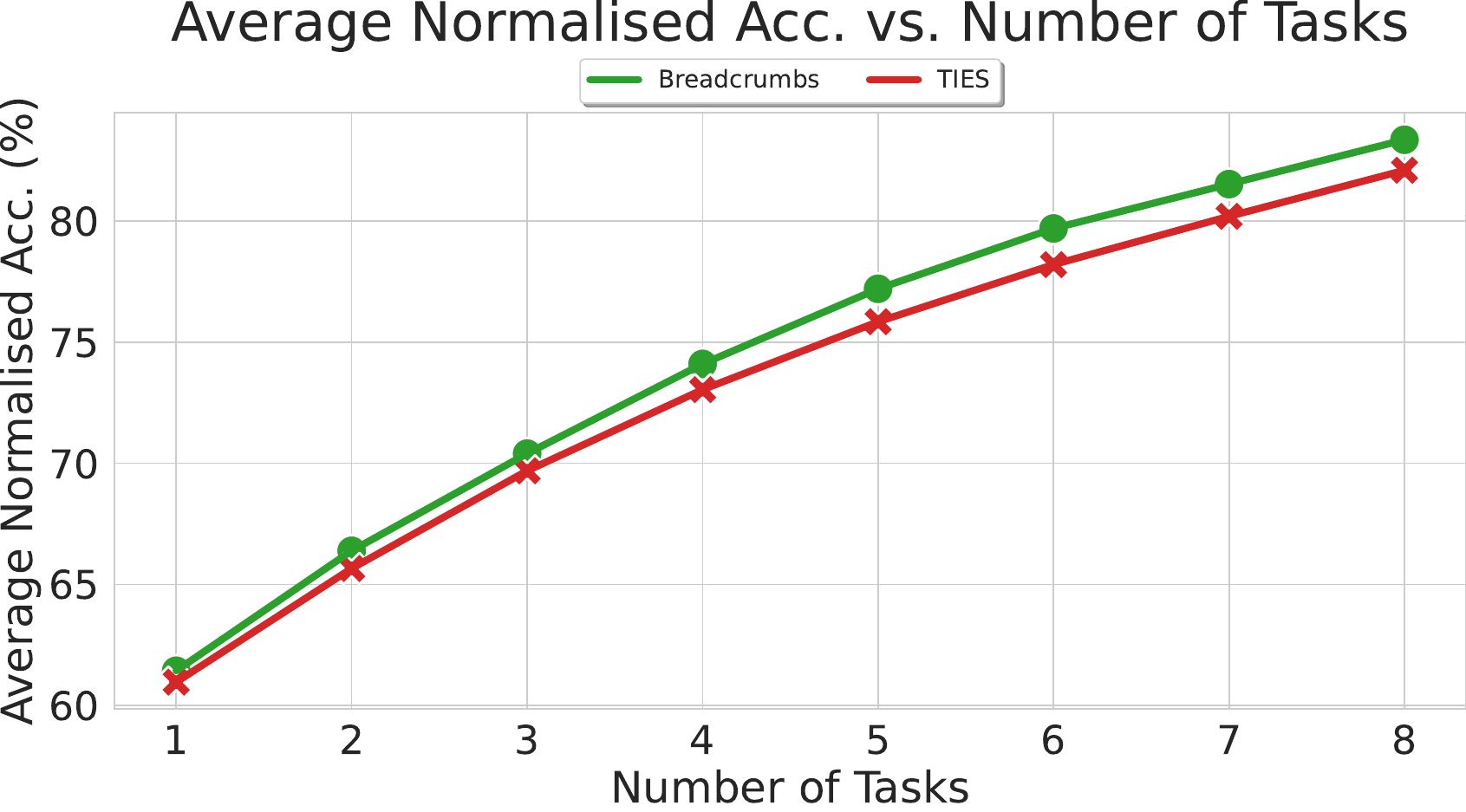}

    \caption{Comparison of Model Breadcrumbs and TIES merging methods across tasks, illustrating Model Breadcrumbs' consistent outperformance, with the performance gap widening as tasks increase. The results underscore the superior practical performance gains of Model Breadcrumbs at scale.}
    \label{fig:ties-comparison}

\end{figure}

In Figure~\ref{fig:ties-comparison}, we compare our method to TIES~\cite{yadav2023ties}, where at each point:
\begin{enumerate*}[label=(\Roman*)]
    \item We show results for \texttt{ViT-B-32} model where we found the best hyper-parameters for that specific number of tasks for each method.
    \item We show the average normalized accuracy over all subsets of the 8 tasks detailed in Section~\ref{sec:data-and-metrics}, amounting to a total of $256=2^8$ combinations.
    \item The evaluation is performed over all 8 tasks at each point.
\end{enumerate*}
As we can see from Figure~\ref{fig:ties-comparison}, Model Breadcrumbs merging consistently outperforms the TIES method at each point, with the performance gap widening as more tasks are considered. This highlights the significant practical performance advantages of Model Breadcrumbs on a larger scale.

\end{document}